\documentclass[11pt]{article}

\usepackage[margin=1in, bottom=1.3in, footskip=0.5in]{geometry}
\usepackage[T1]{fontenc}
\usepackage[utf8]{inputenc}
\usepackage{lmodern}
\usepackage{graphicx}
\usepackage{booktabs}
\usepackage{hyperref}
\usepackage{fancyhdr}

\begin{document}
\begin{center}
{\LARGE\bfseries Closing the AI Trust Gap:\\[0.3em]
The Case for Independent Certification\\[0.3em]
for Trustworthy AI}\\[1.5em]

Trisevgeni Papakonstantinou\textsuperscript{1},
Cansu Canca\textsuperscript{2},
Farah Nanji\textsuperscript{3},
Waheedullah Pardess,
Jen Weedon\textsuperscript{4},
Jasmijn Remmers\textsuperscript{5},
Eliza Krigman\textsuperscript{6},
Matthew Ball,
Yalda Daryani\textsuperscript{7},
Kiran Iqbal\textsuperscript{8},
Francielle Vargas\textsuperscript{9},
María Llorente Sánchez,
Joe Humphreys,
Fendi Tsim\textsuperscript{10},
Kelly Fitzpatrick,
Jeff Dunn,
Catherine Feldman\textsuperscript{11}

\vspace{0.8em}
\small
\textsuperscript{1}University College London\quad
\textsuperscript{2}AI Ethics Lab\quad
\textsuperscript{3}University of Cambridge\quad
\textsuperscript{4}Columbia University\quad
\textsuperscript{5}MATS\\
\textsuperscript{6}Tech with Intention\quad
\textsuperscript{7}University of Southern California\quad
\textsuperscript{8}Kyushu University\quad
\textsuperscript{9}University of Chile\\
\textsuperscript{10}BehSci Meets AI\quad
\textsuperscript{11}Digital Trust Council

\vspace{1em}
\normalsize
\textit{Digital Trust Council}\\
\textit{July 2026}
\end{center}

\vspace{2em}

\begin{abstract}
Over the past decade, responsible AI (RAI) has produced a substantial body of practice for
identifying and mitigating the risks AI poses in high-stakes settings. Yet this work has not
produced a market that rewards trustworthiness. Firms that invest seriously in safety, fairness,
and oversight cannot consistently prove to consumers, regulators, and shareholders that their
systems go beyond the bare minimum of compliance. What is missing is a way for society to
recognize or compare the difference. The result is a trust gap: a structural condition in which
responsible development efforts happen inside organizations but produce no external,
independently recognized and verifiable signal of trustworthy outcomes.

We argue this gap is sustained in part because of a focus on responsible AI (a matter of
internal process) as opposed to trustworthy AI (a matter of independently verifiable real-world
outcomes), and that it persists because of three compounding failures: (1) the market cannot
distinguish trustworthy systems from their imitations; (2) evaluation targets models and
outputs rather than deployed sociotechnical systems and their outcomes; (3) the
measurement ecosystem is oriented toward avoiding harm rather than demonstrating benefit.
Reviewing existing AI governance instruments and comparing them to certification regimes in
healthcare, sustainability, and security, we show that none integrate a governance baseline,
independently verified positive-outcome evidence, and market signaling in a single
framework. We propose independent, outcome-oriented certification as the connective layer
that can close the trust gap, complementing regulation and internal governance by making
trustworthiness measurable, comparable, and commercially rewarded.
\end{abstract}

\section{Introduction}

Over the last decade, the field of responsible AI (RAI) has developed risk management
practices, data and model evaluation guidelines, governance frameworks, and tools to assess
and enhance the safety, fairness, benefit, ethical implications, and overall impact of AI
systems. This field has emerged in response to the increasingly urgent risks AI poses in
high-stakes sectors, such as law enforcement, education, healthcare, and finance. However,
the resulting governance structure has not produced a market that incentivizes doing more
than the bare minimum. Companies that invest substantially in safety, fairness, and oversight
cannot reliably distinguish their products from those that merely claim responsibility.
Consumers and regulators do not have a reliable, accessible, and consistent way to recognize
and compare these underlying practices. Investors lack the instruments to factor them
meaningfully into valuations or procurement decisions. The market currently has no
agreed-upon framework for assessing whether a given AI system, or its integration into a
product or process, does more good than harm. In practice, the AI governance ecosystem has
built a strong floor for harm reduction through compliance, but not a structured way to
recognize or reward systems that go further. The regulatory floor thus risks becoming a
ceiling: organizations optimize for compliance with minimum standards rather than striving to
demonstrate that their systems produce real benefit.

Public trust data reflects this structural deficit. In the United States, a 2025 Pew survey found
that 59\% of the public and 55\% of surveyed AI experts had little or no confidence in companies
to develop and use AI responsibly \cite{Pew2025}. Globally, fewer than half of respondents report they trust
AI systems at all, the majority expressing concern about AI's broader impact on society, such as
cybersecurity risks and the loss of human interaction \cite{Gillespie2025}.

Efforts around RAI have historically focused on the internal practices organizations adopt to
guide how AI is developed and deployed. This body of work has established a shared
vocabulary and advanced methods for identifying and mitigating risks in AI systems, including
the adoption of binding regulation across multiple jurisdictions, among them the EU AI Act and
the emerging state-level frameworks in the United States \cite{EUAIAct2024}. It has not, however, culminated
in a standardized framework that enables consumers, buyers, implementers, and regulators to
understand and compare AI systems on the basis of their actual impact and outcomes
\cite{Birhane2024}. RAI practices, while essential, have remained largely non-standardized and voluntary,
without producing externally verifiable signals of real-world performance and impact. These
externally verifiable markers are crucial because they transform internal practices into
observable evidence that others can scrutinize, compare, and reward. Such signaling, we
argue, requires an ecosystem where standards are clear and agreed-upon. Where standards
are clear, external bodies can assess and compare systems, not only in isolation, but
according to their real-world implications. Facilitating this capability, we argue, is what would
close the gap between responsible AI and trustworthy AI.

Trustworthy AI and responsible AI are often used interchangeably. Yet a conceptual distinction
can be drawn between them—and addressing this distinction is at the crux of our argument
for what constitutes good AI. Responsible AI has focused on implementing risk mitigation
processes and ensuring that systems are designed, developed, and deployed responsibly.
This means engaging with ethical values, legal rules, societal goals, and technological
constraints, and implementing the resulting decisions on the AI system. As Buijsman \cite{Buijsman2024}
argues, a “process-based approach” informs the steps that were taken, “but not what the
outcome is of these steps [and] fails to inform us on both the impact of the decision-making
system and on how successfully citizens’ voice is heard and considered” (p.~34). Trust is an
outcome of a process. If we know, for example, that a car is built with all the best practices,
processes, and systems for safety, we can trust that it is safe and reliable. However, “knowing”
this requires an added layer of infrastructure. We cannot simply take the car dealer’s word for
it; instead we rely on established safety standards, tests, and production certificates that can
be acquired only when the car clears those tests. Our trust depends not just on how the car is
built, or whether it is strictly compliant, but on how it performs—in other words, its real-world
outcomes. We also follow up and take note of incidents once the car model hits the road, until
it retires. Now let us replace the car with an AI system in this analogy.

Trustworthy AI requires the AI system’s real-world impact to be demonstrably and
independently verifiable, producing safe, reliable, fair, and beneficial outcomes for the people
it affects, across deployment contexts and over time. It is a continuous and relational property,
earned through repeated interactions in real-world settings and dependent on iterative
evidence that a system behaves as promised for the communities who rely on it, rather than
on one-off assurance checkpoints or stated intentions \cite{Afroogh2024,Gillis2024,Amugongo2026}. Trustworthy AI is
about whether systems can demonstrate, with evidence the public can inspect, that they meet
agreed standards in practice. This understanding of trustworthy AI broadly aligns with the
NIST framework \cite{NISTAIRMF2023} as well as the European Commission’s Ethics Guidelines for Trustworthy AI
\cite{EUAIAct2024}. But to date, most activity has focused on building and refining responsible AI processes
inside organizations, rather than creating external mechanisms that show if those processes
have produced genuinely trustworthy systems.

Given that most AI systems are opaque \cite{Burrell2016}, internal governance is an essential element of
the governance ecosystem, but alone, it is fragile in practice. Internal governance is
important, but as corporate priorities shift, it can be scaled back or relocated with little visibility
to the outside world. In recent years, Meta has disbanded its Responsible AI team, Microsoft
has eliminated its Ethics \& Society division, and OpenAI repeatedly reorganized its alignment
and safety functions under successive commercial pressures \cite{Statt2023,Clark2023,Feiner2024}. It is a
necessary component of the governance ecosystem, but cannot, on its own, serve as a
reliable signal of trustworthiness to those who have to decide which systems to adopt,
regulate, or rely on. Building on the traditional requirements of trustworthy AI—such as
reliability, security, explainability, privacy, fairness, and accountability—we also argue that in
high-stakes contexts the evidentiary bar for trustworthiness should extend beyond harm
avoidance to include evidence of positive outcomes.

This lack of standardized practices and verifiable outcomes in the current governance
infrastructure has produced what we call the trust gap: a structural condition in which
responsible AI has built a body of knowledge and practices for risk mitigation but has not
produced the communicable, verifiable, and outcome-oriented infrastructure for earning
well-founded trust. In practice, we identify three challenges that reinforce one another in
creating the trust gap: (1) the market cannot reliably distinguish systems that are genuinely
trustworthy from those that merely perform responsibility, (2) evaluation practices focus on
model outputs and performance on benchmarks rather than on the outcomes of the
socio-technical systems involved, and (3) the governance and measurement ecosystem is
oriented toward avoiding harm rather than demonstrating whether deployment produces
positive outcomes that justify the risks involved.

This paper moves through three levels of analysis. It begins in Section~\ref{sec:three-gaps} with
a structural diagnosis of why internal responsible AI activities alone do not automatically
produce public trust. In Section~\ref{sec:standards}, it analyzes a sample of existing governance and
assurance instruments to determine which parts of the assurance problem they already cover
and what gaps remain. In Section~\ref{sec:adjacent}, it looks to established certification regimes in
other high-stakes sectors to identify design patterns for turning complex performance into
market-legible signals. It then explores in Section~\ref{sec:measuring} what a credible assurance
framework would need to do, not only to measure harm reduction but to verify the outcomes
that actually justify deployment. The paper closes with a concrete proposal: that independent,
outcome-oriented certification can serve as the mechanism that turns those claims into a
signal buyers, regulators, and investors can act on.

In brief, the argument is as follows. In complex, high-stakes industries where reliability is
difficult for stakeholders to verify independently, such as finance, healthcare, cybersecurity,
sustainability, and food safety, certification has emerged to bridge information asymmetries
between producers and stakeholders, translating complex technical evaluations into legible
external signals that help consumers, investors, procurement leaders, and regulators
distinguish between organizations that merely claim responsibility and those that can
substantiate it through independent verification. This paper argues that independent,
outcome-oriented certification can serve as a core market mechanism that complements
regulation and internal governance by providing the connective infrastructure through which
trustworthiness becomes accountable, comparable, and commercially meaningful. Our goal is
not to propose another set of principles, but an account of the institutional infrastructure
through which trustworthiness can be measured, verified, and rewarded.

\section{Three Structural Gaps Explaining the Trust Gap}
\label{sec:three-gaps}

Why has a decade of responsible AI activity failed to produce a market that rewards
trustworthiness? The answer lies in three interlocking structural gaps, which this section
examines. The market cannot distinguish responsible AI from its imitation, and so does not
reward it (Section~\ref{sec:market-gap}). The evaluation methods meant to supply that distinction target the model and
its outputs in isolation rather than the real-world outcomes in which trust is actually formed
(Section~\ref{sec:evaluation-gap}). And the measurement ecosystem is oriented almost entirely toward the harms AI should
avoid, with no equivalent infrastructure for the benefits it should produce (Section~\ref{sec:benefit-gap}). These are not
independent problems but mutually reinforcing ones: the market cannot reward what the
measurement ecosystem cannot demonstrate, and the measurement ecosystem cannot
demonstrate what it does not know how to measure. Addressing them, we argue, requires not
another framework but a connective layer that ties existing instruments into a single, verifiable
signal.

\subsection{The Market Cannot Distinguish and Reward Responsible AI}
\label{sec:market-gap}

The current market's rewards are misaligned with creating trust. Two dynamics, working
together, make robust responsible AI commercially unviable for most firms: incentives reward
speed and engagement rather than trustworthiness, and investment in responsible AI
practices often does not yield a corresponding return, even from investors who say they value
it.

\textit{Incentives reward speed and engagement, not trustworthiness.} Companies are pushed to
reach the market as quickly as possible, and once a system is deployed, user engagement
rather than demonstrated trustworthiness is what gets rewarded. The trajectory of social
media over the past two decades shows where this leads. Public trust has eroded to the point
that Australia has now barred access for under-16s, recently followed by the UK, with more
than ten other jurisdictions weighing similar restrictions~\cite{AustralianGov2025,UKGov2026,CEPA2026}. This trajectory was driven primarily by a business model that
rewarded engagement at all costs, incentivizing design patterns that maximized profit through
outrage, addiction, and misinformation despite repeated ethical commitments~\cite{Wu2016,HumanRightsWatch2019}. The deeper problem was not that engagement became corrupted
over time but that it was never a good proxy for wellbeing to begin with. Engagement tracks
attention and revenue, not user benefit, wellbeing, or durable trust~\cite{Docherty2022}.
When the metric that drives revenue is decoupled from wellbeing, doing right by users
becomes a competitive disadvantage.

AI is reproducing this pattern across layers. Frontier labs move quickly, do the minimum
needed to avoid major safety setbacks, and stay globally competitive. At the product layer, the
same engagement logic takes hold: conversational systems are optimized for stickiness, and
documented tendencies toward sycophancy, telling users what they want to hear, improve
satisfaction metrics while degrading reliability, in some cases to the point of offering guidance
for self-harm if requested~\cite{Sharma2023,SchoeneCanca2025}. Companies that
build on or procure these systems face the same pressure to adopt AI to satisfy shareholders
and stay competitive, without an equivalent incentive to ensure it is trustworthy.

\textit{Investment in responsible AI does not yield a corresponding return.} Surveys show that firms
investing in responsible AI and governance can adopt AI faster (Bain; IBM). This is a major
business case for a wide-scale and robust responsible AI implementation. But the upfront cost
is high, governance can slow deployment relative to ``move fast and break things'',
and, most
importantly, these efforts remain invisible to the external stakeholders who would translate
them into value: consumers do not know, and investors cannot tell. The result is what Akerlof
called a ``market for lemons''~\cite{Akerlof1970}:
when buyers cannot reliably distinguish
high-quality goods
--- here, AI systems that demonstrably produce good outcomes ---
from
low-quality ones wrapped in identical marketing language, they discount all claims equally. A
firm that absorbs the costs of safety teams, ethics boards, red-teaming, and compliance
cannot command a premium for them, and the business case for investment remains weak at
best.

This is clearest among investors, who believe responsible AI is good business yet cannot
appraise it. In a 2026 survey of 56 venture-capital practitioners, 73\% agreed that companies
with stronger responsible AI practices are likely to be more financially successful, both by
avoiding costly incidents and by earning the customer trust adoption requires, yet only 27\%
felt they had internal expertise on the subject, and just 14\% rated their own risk-assessment
capabilities as good~\cite{NixonVromen2026}. Broader surveys find the same gap between
conviction and capability~\cite{PwC2025}. What cannot be reliably evaluated is difficult to price,
and what is not priced is rarely rewarded.

One might think that transparency dissolves this information asymmetry: if firms disclosed
how their systems behave, buyers could reward the trustworthy ones. But disclosure proves
to be a double-edged sword, because its costs and benefits fall asymmetrically. A firm that
documents its systems'
limitations through model or system cards~\cite{Mitchell2019}
effectively hands a roadmap of its weaknesses to regulators, plaintiffs, and competitors,
where a documented, unaddressed risk can be more damaging than one never
acknowledged; a competitor that says little incurs none of these costs~\cite{Sadek2024}.
And because voluntary disclosures are unverified, they generate only limited trust in any case.

Absent some external way to generate credible signals, AI is positioned to repeat social
media's sequence, in which harms accumulate first and governance catches up later.
Regulation does arrive---the EU's Digital Services Act is one example---but typically only after
significant damage is done. A better market design would reverse this order. Reliable signals
of trustworthy performance would let investors, buyers, and users favor systems that are
safer and more beneficial, turning responsible AI from a cost center into a competitive
advantage. Producing such signals before harm accrues is the function we argue certification
can serve.

\subsection{Evaluation Targets the Model and its Outputs, Not its Real-World Outcomes}
\label{sec:evaluation-gap}

If certification is to generate a credible signal of trustworthiness, it must rest on evaluation
methods capable of measuring what trustworthiness requires
--- and current established
practices cannot. The assurance toolkit was built to answer a narrower question, how a model
performs on a battery of pre-release tests, rather than the one trustworthiness needs, and the
gap is structural rather than incidental. Four limitations are particularly consequential: (1) the
unit of evaluation is the model rather than the system; (2) model evaluations are increasingly
gameable by the very systems that most warrant scrutiny; (3) post-deployment monitoring is
largely absent; and (4) existing standards address only parts of the evaluation lifecycle,
without grounding it in the sociotechnical setting or enforcing it against the deployment
contexts where trust is actually formed.

The unit of evaluation in current AI assurance is, almost everywhere, the model outputs rather
than outcomes. Outputs are properties of the model or system in controlled settings:
benchmark scores, refusal rates, fairness metrics, red-team results. They matter, but they are
not the same as outcomes: the real-world effects of deployment on people, institutions, and
environments over time. A system can perform well on static tests yet still fail to deliver
meaningful benefit, or even cause subtle, cumulative harm once embedded in real workflows
and existing societal structures.

Trust cannot be based on a model in isolation; its object is, rather, sociotechnical,
system-level real-world outcomes that depend on deployment context, access controls,
organizational processes, and the safeguards around the model---none of which
output-focused, benchmark-style model evaluations can fully capture~\cite{Weidinger2023}. Even setting aside the strategic behavior discussed below, model-level evaluation
misses the context that determines real-world outcomes, even in traditional AI:
who uses the
system, under what conditions, for what purpose, with what oversight, and exposed to what
populations. These details determine which risk profiles apply and which benefits actually
accrue, yet they sit outside the boundary of what a model evaluation actually measures.

A second problem is that the evaluations can be increasingly unreliable. As frontier systems
grow more capable, they become increasingly able to detect when they are being tested and
adjust their behavior accordingly. METR, an independent AI evaluation organization, has been
explicit about the implication that pre-deployment capability testing alone is no longer a
sufficient accountability strategy. Evaluations risk measuring how a model wants to appear to
behave, not how it will behave under real deployment conditions. Benchmark-based
evaluation can thus systematically underestimate risk for precisely the systems that warrant
scrutiny, and may be directionally misleading rather than merely imprecise.

A third problem is temporal. Even a flawless pre-deployment evaluation captures only a single
moment in a system's life, yet trustworthiness is a continuous property. Models drift as inputs
and usage shift, users apply systems in unanticipated ways, and many harms surface only in
interaction. The assurance ecosystem is built almost entirely around pre-release checkpoints,
with little infrastructure to observe systems in use. The contrast with other high-stakes
sectors is stark---aviation has mandatory incident reporting through the NTSB, and
pharmaceuticals have post-market adverse-event surveillance---whereas AI is only
beginning to acquire comparable infrastructure. In the US there is still no standardized,
mandatory channel for reporting failures or general requirement that deployers monitor
systems against their approval conditions. In the EU, the AI Act introduces serious-incident
reporting and post-market monitoring obligations for high-risk systems (Articles~72, 73) and
deployers (Article~26), which are already in force for systemic-risk GPAI models and will apply
to high-risk systems from August 2026. Though enforcement maturity, coverage, and a
centralized incident database still lag compared to the aviation and pharmaceutical regimes.
Voluntary efforts such as the AI Incident Database document the need but lack the authority
and coverage to meet it~\cite{McGregor2021}.

Existing frameworks recognize parts of these problems. NIST's AI Risk Management
Framework bridges pre-deployment and operational monitoring through its Test, Evaluation,
Validation, and Verification (TEVV) function, reflecting the view that risk measurement should
be continuous rather than one-off, and related standards work, including ISO/IEC~42119, is
extending the technical foundation for lifecycle testing~\cite{NISTAIRMF2023}. But two gaps remain. First, the
lifecycle gap that Raji et al.~\cite{Raji2020} identified more than five years ago
--- that available
auditing tools tend to cover discrete stages rather than the end-to-end system pipeline
--- has
not closed. A 2025 study of the AI audit-tooling landscape, drawing on interviews with 35
practitioners and an analysis of 435 tools, finds that while many support evaluation and
standard-setting, they fall short of the institutional infrastructure needed to turn audit findings
into meaningful accountability outcomes~\cite{Ojewale2025}. The instruction to evaluate
across the lifecycle exists; the infrastructure to enable it at scale or enforce it does not.
Second, these frameworks focus on technical performance and risk measurement, whereas
trustworthiness requires demonstrating safe, effective, and beneficial performance in context,
accounting for the people involved, the deployment setting, and the outcomes for affected
users.

Taken together, these four gaps describe an assurance regime that measures the wrong unit,
with instruments the most capable systems can game, at a single point in a system's life,
against standards that stop short of the deployment context where trust is actually formed.
Measurement matters only if it captures the full sociotechnical system; without system-level
verification, the market cannot trust the signal even when the underlying metric is sound.

\subsection{Responsible AI is Oriented Towards Harm Avoidance, Not Benefit}
\label{sec:benefit-gap}

The incentive problem is compounded by the fact that the current measurement ecosystem is
far better at tracking what AI should avoid than what it should produce. When AI systems are
evaluated, the ecosystem structurally focuses on accepting a minimum threshold of risk and
mitigating it, or on what systems should not do. As Saari and Mügge~\cite{SaariMuegge2026} argue, risk
management frameworks narrow governance attention toward harms that can be clearly
defined, measured, and mitigated, while broader questions about social value, distributional
impacts, and public benefit often remain outside their scope. As a result, once organizations
satisfy compliance requirements and demonstrate that risks have been reduced to acceptable
levels, there is little institutional pressure to show that AI systems generate outcomes that
justify their deployment. We have no comparably mature, shared infrastructure for measuring
positive outcomes, or what they should deliver.

The tools for measuring risk are growing across industry, academia, and civil society:
Stanford's HELM Safety, Microsoft's Fairlearn, the Future of Life Institute's AI Safety Index, and
MIT's AI Risk Repository, among others~\cite{StanfordHELMSafety2024,FairlearnGuide,FutureOfLife2025,MITRiskRepo}. All of these measure what to avoid. That
orientation has created a floor-as-ceiling paradox: most governance tools are designed to
raise and enforce a minimum standard, avoiding certain failure modes, but in practice that
minimum is treated as the end goal. As long as systems clear the floor, there is little incentive
to ask whether they generate value commensurate with their risks. If measurement stops at
risk avoidance, the market can reward only compliance and harm reduction, not systems that
actually create benefit.

The benefit side has no equivalent maturity. There is no mature, widely adopted metric
taxonomy for whether an AI system actually delivers value to the people it affects.
Outcome-level questions---has access improved? Has equity improved? Have outcomes
improved for the populations the system was deployed to serve?---are addressed in case
studies and academic literature, but they are absent from the major guidelines, frameworks,
and certifications that shape how organizations behave at scale~\cite{NISTMonitoring2026}. While these
documents affirm the importance of benefit, they largely lack standardized, enforceable
metrics or evidentiary thresholds for positive outcomes in deployment.

The asymmetry reflects how the accountability ecosystem was built around harm prevention
and process auditing. When the entire assurance environment orients around demonstrating
compliance and conformity, rather than generating evidence of benefit, the result is a field in
which AI systems can meet baseline safety requirements while still producing adverse
outcomes over extended use, in ways that might be subtle and hard to identify. Even where a
system can be shown not to be actively harmful, that floor is not enough to motivate
deploying a technology with documented risks when the potential upside is only neutral. To
make the risk trade-off rational for individuals, organizations, and society, one must be able to
demonstrate a markedly beneficial upside. That is why positive-outcome measurement
matters. Recent work on positive alignment frames such outcomes as conditions for AI to
support human flourishing, not merely to avoid undermining it~\cite{Laukkonen2026}, and
efforts such as the IEEE CertifAIEd joint specification for trustworthy AI systems~\cite{IEEECertifAIEd2023}
represent progress toward operationalizing positive impact within a standards framework,
though questions remain about adoption incentives and feasibility at scale.

Taken individually, each of these challenges has been identified before. They are not novel
observations, and serious work is underway on each. But they cannot be solved separately,
because each one reinforces the others: the market does not reward trustworthiness in part
because the measurement ecosystem cannot demonstrate it; the measurement ecosystem
cannot demonstrate it in part because what is measured is often the model in isolation, not the
full system in which trust forms; and the market cannot reward better performance when
there is no comparable signal predictive of real-world outcomes. A solution that addresses
any one challenge in isolation runs into the limits imposed by the others. Better fairness
metrics do not change firm behavior if there is no commercial reward for using them. Lifecycle
evaluation does not produce trust if its results cannot be compared across firms. And market
signals do not work if what they signal is not predictive of real-world outcomes. What is
needed, thereby, is a connective layer, not an additional framework. The field does not suffer
from a lack of instruments or well-developed frameworks; the obstacle is their underlying
assumptions and surrounding ecosystem, which prevent them from producing AI that is
trustworthy, outcome-oriented, and benefit-driven. A connective infrastructure that integrates
existing instruments into a single market-legible signal, enabling lifecycle assessment,
positive-outcome measurement, and independent verification, has the potential to shift the
incentive structure needed to address these challenges. Independent auditing and
certification can be that mechanism.

\section{Evaluating Current Standards: Why the Trust Gap Remains}
\label{sec:standards}

So far we have argued that trustworthy AI cannot be reduced to more elaborate processes or
more frequent self-assessment alone. The core issue is not the absence of best practices, but
the absence of a mechanism that turns governance, measurement, and verification into a
signal the market can recognize---one that incentivizes companies to do more than the bare
minimum. This section examines whether existing AI governance and assurance instruments
already solve that problem, and where their coverage is strong or weak. We review the most
relevant frameworks against a set of dimensions drawn from the conformity-assessment
literature (ISO/IEC 17000:2020) and algorithmic-auditing scholarship~\cite{Raji2020}, relying
on desk research and structured comparative assessment across primary documentation. We
do not endorse a specific scheme, and we ground every gap we identify in publicly available
sources. By mapping the current structures and coverage areas and gaps, we can better
understand the best path towards fulfilling the design requirements for trustworthy AI systems
and credible market signals.

\subsection{Mapping the Landscape}

This study selects ten current AI governance instruments for structural diversity rather than
exhaustiveness, taking one representative example from each major category:
intergovernmental guidance, binding regulation, management-system standards, technical
benchmarks, national assessment tools, professional certifications, consumer-facing labels,
and corporate responsible AI toolchains. Each instrument is scored independently against
eight pre-specified dimensions, documenting confidence levels for every score. The
dimensions are grounded in the conformity-assessment vocabulary of ISO/IEC 17000:2020,
empirical audit work such as Raji et al.~\cite{Raji2020}, and cross-sector certification analogues
including LEED, ISO 27001, IEC 62443, and SOC 2.

\begin{table}[t]
  \centering
  \small
  \caption{Comparative assessment of current AI governance instruments}
  \label{tab:governance-instruments}
  \begin{tabular}{lccccccc}
    \toprule
    Instrument & Impact & Meas. & Assur. & Lifecycle & Mkt Sig. & Cred. & Legit. \\
    \midrule
    UNESCO + RAM/EIA         & Partial & Strong  & Partial & Partial & None   & Strong & Strong \\
    OECD AI Principles + DDG & Partial & Partial & Partial & Partial & None   & Strong & Partial \\
    NIST AI RMF              & Partial & Strong  & Partial & Strong  & None   & Strong & Partial \\
    EU AI Act (high-risk)    & Partial & Partial & Strong  & Strong  & Partial & Strong & Partial \\
    ISO/IEC 42001            & None    & Partial & Strong  & Strong  & None   & Strong & Partial \\
    HELM Safety benchmark    & None    & Strong  & Partial & Partial & None   & Partial & None   \\
    Canada AIA               & Partial & Strong  & Partial & Partial & Partial & Strong & Partial \\
    IEEE CertifAIEd          & Partial & Strong  & Strong  & Partial & Partial & Strong & Partial \\
    Digital Trust Label      & Partial & Strong  & Strong  & Partial & Strong  & Strong & Partial \\
    Corporate RAI toolchains & Partial & Strong  & None    & Partial & None   & Partial & None   \\
    \bottomrule
  \end{tabular}
\end{table}
\noindent
\textit{Strong} (green) \textit{Partial} (amber) \textit{None} (red). Full scoring rationale: Annex~A (companion document).
Sources: EU AI Act (2024); ISO/IEC 42001 (2023); NIST AI RMF (2023); OECD AI Principles and Due Diligence Guidance (2026);
UNESCO Recommendation and RAM/EIA (2021/2024); HELM Safety, Stanford CRFM (2024); Canada AIA, Treasury Board (2023);
IEEE CertifAIEd (2023); Digital Trust Label, SGS/SDI (2024); Microsoft Responsible AI Toolbox (2024).

\subsection{Findings}

The assessment reveals a consistent pattern. The instruments do well at what they were
designed to do, which is process governance and risk management, but leave out the
dimensions that matter the most for a credible market signal.

The current landscape is strong on process governance and risk management. ISO/IEC
42001, the EU AI Act's high-risk conformity regime, and the NIST AI RMF collectively provide a
good foundation for examining that an organization has the right structures, processes, and
controls in place. An organization that satisfies all three has done significant governance
work. That is a usable baseline, and it is where any new certification architecture should start,
not where it should end.

In these frameworks, outcome assurance is absent. None of the instruments assessed
requires evidence that an AI system is actually producing benefit for the communities it
serves. This is not a flaw in the individual instruments; most were built for risk management or
regulatory compliance, and they do that competently. The gap is that the architecture as a
whole has no layer for demonstrating positive outcomes. A system can satisfy every
requirement of ISO/IEC 42001, pass the EU AI Act's high-risk conformity assessment, and
align fully with NIST AI RMF, yet still deliver a product of no measurable benefit to anyone who
uses it.

Coverage across the lifecycle is similarly uneven. The NIST AI RMF's TEVV function is the
most explicit existing treatment of continuous, lifecycle-spanning evaluation, but it is
guidance, not an enforceable standard. Most instruments concentrate either on
pre-deployment governance documentation or on organizational management systems.
Deployment context, ongoing monitoring, and post-market accountability are addressed
inconsistently across the landscape, and verified almost nowhere.

Stakeholder legitimacy is also limited. The populations most affected by AI deployment have
the least voice in any of the governance instruments assessed here. UNESCO's participatory
EIA methodology is the only exception, and even that stops short of translating community
input into outcome metrics that certification could operationalize. Community involvement in
defining what benefit means is a substantive requirement, not a procedural one. For a
certification framework that aims to be substantively legitimate, a system cannot credibly be
certified as serving its communities without established processes of active involvement and
co-production through which those communities define what benefit means for them. A few
initiatives sit closer to the target such as ForHumanity, the RAI Institute, and IEEE CertifAIEd.

And most critically, market signaling is almost entirely absent. The AI trust market is not
failing because assessment tools do not exist; it is failing because their outputs do not
translate into signals that buyers, regulators, and the public can act on. Most current
instruments are internal governance tools or compliance credentials that do not translate
technical performance into a signal those stakeholders can readily use. In practice, this means
that organizations can invest in responsible AI without gaining a clear competitive return for it.
The only outlier that generates a recognized consumer-facing trust mark with documented
procurement value is the Digital Trust Label, a certification scheme developed by the Swiss
Digital Initiative (SDI). Close neighbors such as IEEE CertifAIEd move toward structured
third-party assurance. But none of these schemes, on their own, integrates a governance
baseline, enforceable risk controls, independently verified positive-outcome evidence, and
tiered market signaling within a single framework. Existing instruments supply important
components, but not yet the full connective layer.

Various frameworks have built important pieces of assurance infrastructure. However, none
combines a governance baseline, a risk floor, independently verified positive-outcome
evidence, and market signaling in a single tiered framework. That is the integration problem
any new framework must solve.

\section{Lessons from Adjacent Sectors}
\label{sec:adjacent}

Three cross-sector comparisons sharpen the design requirements for market signaling and
benefit measurement in AI certification: healthcare, sustainability, and security.

Healthcare has the most developed infrastructure for attributing outcomes to interventions
under conditions of causal complexity. Its clinical evidence hierarchy (from expert opinion
through observational studies to randomized trials and systematic reviews) provides the
template for an evidence-maturity model. The lesson for AI is not that systems must meet
clinical-trial standards, but that tiered evidential expectations with explicit maturity levels are
operationally workable: they create progressive incentives to invest in stronger evidence over
time, rather than treating all evidence as equivalent or none as sufficient.

LEED (Leadership in Energy and Environmental Design) is the world's most widely used green
building rating system. It shows how a multi-dimensional benefit framework can be translated
into a tiered certification that non-experts can navigate and act on. Its partial-credit
architecture across energy, water, materials, indoor environment, and site sustainability allows
buildings to earn certification at four levels without requiring perfect performance on any
single criterion. The AI equivalent is a system that excels on safety and transparency but has
not yet generated longitudinal evidence to earn an entry-tier certification, with clear direction
toward the remaining evidential gaps and a commercial incentive to close them.

SOC~2 (System and Organization Controls 2) is an independent auditing standard that
evaluates how securely a service organization manages customer data to protect privacy and
confidentiality. It shows how a voluntary framework becomes effectively mandatory through
procurement leverage rather than legislative mandate. Enterprise buyers began requiring
independent security attestation as a vendor-selection criterion, and adoption accelerated
through supply-chain requirements without regulatory enforcement. The same logic applies to
trustworthy AI certification: if it becomes a meaningful differentiator in procurement,
particularly in enterprise and public-sector markets, where AI risk assessments are already
cumbersome and non-standardized, adoption can follow commercial interest without waiting
for a regulatory mandate.

The importance of demonstrating benefit is not in dispute; every governance framework
reviewed above affirms it in some form. The harder question is whether benefit evidence,
independently assessed and tied to certification tiers, can convert trustworthiness from a
stated priority into a market advantage. The experience of healthcare, sustainability, and
security, among others suggests it can, but only once the infrastructure that makes
trustworthiness visible and comparable exists. The analysis herein confirms the structural
pattern outlined in Section~\ref{sec:three-gaps}:
current instruments work hard to strengthen the risk floor and
assess outputs, but offer little shared infrastructure for assessing outcomes or rewarding
systems that demonstrate value above the minimum.

\section{Measuring and Certifying Trustworthiness}
\label{sec:measuring}

The landscape analysis in Section~\ref{sec:standards} shows that existing instruments cover important parts of
the assurance problem, but none satisfies all the conditions a trustworthy AI framework needs
to close the trust gap laid out in Section~\ref{sec:three-gaps}. Closing that gap calls for an accountability
infrastructure that integrates several complementary requirements, which together make
trustworthiness establishable at the system level: it must evaluate the right component, across
the right stages, independently, and in a way that can affect decisions. Crucially, this does not
depend on inventing new governance tools. Most of the building blocks already exist; leading
organizations conduct extensive evaluations, including adversarial tests, invest in red-teaming
models, maintain monitoring systems, and produce governance artifacts as part of standard
development and deployment~\cite{FutureOfLife2025}. So the task is to organize these
existing practices, such as internal risk registers, model and system documentation, safety
evaluations, and incident reports, into a coherent structure that independent third parties can
verify and compare. The framework must satisfy five design requirements:

\textit{Produce standardized, comparable results.} The output of an assurance framework must
enable consumers, governments, and enterprise buyers to rank systems against one another
on their trustworthiness. That is possible only if the framework delivers standardized,
comparable results.

\textit{Evaluate deployed systems, not just models.} Evaluation must span the system's lifecycle,
including how models interact with interfaces, organizational policies, and real users in the
wild. It must be ongoing and grounded in evidence of real-world performance, not just
compliance at the point of release. To evaluate deployed systems rather than isolated models,
assurance must look beyond model outputs to the sociotechnical system around them. In
practice, this means examining how data is collected and sourced; how models are trained
and aligned; how they are configured and used in specific deployment contexts; how
performance and incidents are monitored over time; how post-deployment auditing, logging,
and redress work; and how organizational governance and policy shape the decisions taken
on the basis of model outputs~\cite{Raji2020}.

\textit{Measure positive outcomes alongside risk.} The framework must develop shared metrics for
benefit delivery, based on research on wellbeing, public-interest AI, prosocial design, and
health outcomes. This body of knowledge exists; what is missing is their integration into
instruments practical enough to apply across deployment contexts. When the assurance
ecosystem is built around harm avoidance, organizations invest accordingly: they document
risks, run benchmarks, and produce artifacts demonstrating that nothing has gone wrong. The
question that matters for deployment---whether something has gone right, whether the system
improved access, reduced unfairness, or justified the trust placed in it---goes unanswered,
because nothing requires answering it. Adding benefit evidence as a condition of higher tiers
does more than reward good actors: it changes what organizations measure, since
organizations invest in what gets measured.

\textit{Be independently governed.} A third-party institution, with no conflict of interest with the AI
industry, must perform the evaluation. It is important for AI model developers to test their own
models for safety and performance, but they cannot earn trust through self-attestation. There
must be clear separation between those who build systems and those who evaluate them.
Because certification bodies can themselves become vulnerable to industry capture, a
credible scheme also requires public-interest governance, transparency about methods and
conflicts of interest, independent oversight, and periodic review of inter-certifier consistency.

\textit{Generate market-legible signals.} The framework should convert complex technical
performance into something that markets can use: an indicator of how trustworthy an AI
model or system is to enable companies to compete on trustworthy capabilities.

A trustworthy AI assurance framework is not economically consequential simply because it is
technically sound. For certification to matter in practice, it must be embedded in a wider set of
market and institutional incentives and able to meaningfully affect decisions. On the supply
side, developers and deployers have no standardized way to demonstrate trustworthy
performance across the lifecycle; on the demand side, buyers, regulators, and investors
cannot consistently reward trustworthiness. The value of independent certification therefore
depends on whether the surrounding institutions can price the signal into commercial choices.

Two objections bear directly on whether certification is the right response to address the trust
gap. One concerns feasibility: certification may demand a measurement infrastructure
organizations do not yet have. The second is legal: regulation, not voluntary certification, may
be the more legitimate instrument. Neither objection defeats certification, but together they
define the conditions it must satisfy.

The feasibility objection overstates what certification requires from scratch. Frontier labs
already produce substantial relevant evidence: pre-deployment evaluations, monitoring logs,
red-teaming outputs, governance documentation, incident records, and compliance artifacts.
The task is not to build new instrumentation but to standardize and independently verify
evidence that already exists, then package it into a form comparable across firms and usable
in procurement, regulation, and investment. A tiered structure distributes the cost in
proportion to ambition: the base tier relies on evidence many responsible organizations
already generate, while higher tiers require progressively stronger outcome evidence. The
framework is also built to evolve. Today's measurement tools set the feasibility floor---what can
be evaluated now, at scale. As positive-impact metrics mature and independent audit capacity
grows, the thresholds for each tier will be revised through structured update cycles.

The legal objection misreads the relationship between the two instruments. Law has greater
enforcement power, democratic authority, and coercive force than voluntary assurance.
However, in fast-moving AI markets, regulation sets the legal floor while often moving too
slowly to shape incentives in real time; by the time a statute is in place, the market may
already have normalized the next generation of systems and risks. Certification can operate
while the regulatory process catches up, giving buyers and regulators a usable signal now
rather than the promise of one later. Legislative action in the United States offers early
evidence that policymakers recognize this need: in June, Connecticut Governor Ned Lamont
signed into law HB~5222, a bill that creates a pilot program for Independent Verification
Organizations to assess whether AI systems meet state-defined safety and governance
objectives~\cite{Connecticut2026,Fathom2026,Virginia2026}.

Certification is therefore not a replacement for regulation but one layer of a three-layer
architecture: regulation sets the legal floor, standards translate that floor into operational and
technical specifications, and certification provides independent verification against those
standards. The layers are complementary.

Perhaps the most practical objection is that certification may not change behavior if buyers do
not care enough. In theory, a credible signal should help procurement, investment, and
regulation distinguish better systems from merely compliant ones; in practice, many decisions
are still driven by price, convenience, speed, or brand. That is exactly why the framework
must be embedded where decisions are actually made: procurement rules, investor due
diligence, enterprise risk processes, insurance pricing, and regulatory recognition. A signal
that does not enter those gatekeeping systems remains informationally interesting but
commercially weak.

The implication is that certification is not a standalone solution but a connective layer, and its
effectiveness depends on two tracks moving in parallel. On the supply side, the field needs an
assurance instrument that can verify trustworthy performance across the lifecycle. On the
demand side, buyers, regulators, and investors must treat that signal as decision-relevant. In
other sectors, certification becomes consequential not because the label exists in isolation but
because surrounding market and policy conditions give it force. Each of these mechanisms
solves part of the problem: regulation creates the floor, self-governance builds internal
capability, and market pressure can reward trustworthiness once it is visible and comparable,
without the downsides of transparency alone. Certification can accomplish this.

\section{Conclusion}
\label{sec:conclusion}

Our goal in this paper has been to show that a market-legible form of trustworthy AI is not
automatically produced by the existing practices of responsible AI. It requires a connective
infrastructure to turn internal processes into independently verifiable real-world outcomes.
We have argued that this is a structural gap rather than a failure of effort: firms compete on
what markets can see, and the market cannot currently see, compare, or reward
trustworthiness. The gap rests on a distinction between responsible AI, a matter of internal
process, and trustworthy AI, a matter of independently verifiable real-world outcomes, and it
is held open by three compounding failures: the market cannot distinguish trustworthy
systems from their imitations, evaluation targets the model rather than the deployed
sociotechnical system, and the measurement ecosystem is built to track harm avoided rather
than benefit delivered. Existing instruments such as ISO/IEC 42001, the EU AI Act, and the
NIST AI RMF build a credible floor but do not assemble into a signal, leaving outcome
assurance thin and market legibility weak. Adjacent sectors---healthcare, LEED, SOC~2---
show that tiered, independently verified signals become consequential only once they are
embedded in procurement, investment, and regulation. Closing the gap therefore calls not for
another framework but for a connective layer: independent, outcome-oriented certification
that evaluates deployed systems, measures benefit alongside risk, produces comparable
results, is independently governed, and generates market-legible signals.

What follows is a choice about sequence. AI can repeat the social-media pattern, in which
harms accumulate first and accountability arrives afterward, or it can build the connective
layer now, while the market is still taking shape. The instruments largely exist; what is missing
is the architecture that makes their results comparable and rewardable. Building it is neither a
purely technical nor a purely regulatory task but an institutional one---and the window in
which it can be done, before the next generation of systems is normalized, is narrower than it
looks.
\bibliographystyle{plain}
\bibliography{references}

\end{document}